\documentclass[runningheads]{llncs}

\usepackage[year=2026,ID=2186]{eccv}

\usepackage{eccvabbrv}

\usepackage{graphicx}
\usepackage{booktabs}
\usepackage{multirow}

\usepackage[accsupp]{axessibility}  

\usepackage{hyperref}

\usepackage{orcidlink}

\begin{document}

\title{FAST3DIS: Feed-forward Anchored Scene Transformer for 3D Instance Segmentation} 

\titlerunning{Feed-forward Anchored Scene Transformer for 3D Instance Segmentation}

\author{Changyang Li \and Xueqing Huang \and
Shin-Fang Chng \and \\ Huangying Zhan \and  Qingan Yan \and Yi Xu}

\authorrunning{Li et al.}

\institute{Goertek Alpha Labs \\
\email{\{first.last\}@goertekusa.com}}

\maketitle

\begin{abstract}
While recent feed-forward 3D reconstruction models provide a strong geometric foundation for scene understanding, extending them to 3D instance segmentation typically relies on a disjointed post-hoc clustering paradigm. Grouping dense pixel-wise embeddings via non-differentiable clustering scales poorly with the number of views and disconnects representation learning from the final segmentation objective. In this paper, we present a \textbf{F}eed-forward \textbf{A}nchored \textbf{S}cene \textbf{T}ransformer for \textbf{3D} \textbf{I}nstance \textbf{S}egmentation (\textbf{FAST3DIS}), an end-to-end approach that effectively bypasses post-hoc clustering. Built upon a foundational depth backbone, our method is efficiently adapted to learn instance-specific semantics while retaining its zero-shot geometric priors. We formulate a learned 3D anchor generator coupled with an anchor-sampling cross-attention mechanism for view-consistent 3D instance segmentation. By projecting 3D object queries directly into multi-view feature maps, our method samples context efficiently. Furthermore, we introduce a dual-level regularization strategy, that couples multi-view contrastive learning with a dynamically scheduled spatial overlap penalty to explicitly prevent query collisions and ensure precise instance boundaries. Experiments on complex indoor 3D datasets demonstrate that our approach achieves competitive segmentation accuracy with significantly improved memory scalability and inference speed over state-of-the-art clustering-based methods.

\keywords{3D reconstruction, 3D instance segmentation, multi-view attention, contrastive learning}
\end{abstract}

\section{Introduction}

Holistic 3D scene understanding is a fundamental task of next-generation spatial intelligence, enabling applications in augmented reality, robotics, and autonomous navigation to perceive the world not just as a collection of pixels, but as distinct, meaningful entities. While traditional approaches treated geometric reconstruction (e.g., SLAM, MVS) and semantic understanding as separate, sequential pipelines, the field has recently transitioned toward feed-forward models~\cite{keetha2026mapanything,koch2025unified,li2025iggt,wang2024dust3r,wang2025pi,Yang_2025_Fast3R,depthanything3,tang2025mv,wang2025vggt}. These methods have demonstrated remarkable capabilities in recovering dense 3D geometry directly from unposed images.

Building upon these geometric foundations, approaches like PanSt3R~\cite{zust2025panst3r} have achieved feed-forward multi-view panoptic segmentation. In such setups, adjacent objects of different categories can be naturally separated relying on semantic labels. However, in the more challenging setting of class-agnostic instance segmentation, where the model must separate objects purely based on physical structure and spatial separation, explicit feature-space regularization is required. Recent works like IGGT~\cite{li2025iggt} and UNITE~\cite{koch2025unified} have attempted to integrate instance-level understanding into the reconstruction pipeline. However, these methods rely on a disjointed post-hoc clustering paradigm: they predict dense, high-dimensional embeddings for every pixel and subsequently employ heuristic, non-differentiable clustering algorithms (e.g., HDBSCAN~\cite{mcinnes2017accelerated}) to group these embeddings into discrete objects. While effective in certain scenarios, this approach suffers from severe scalability bottlenecks. The clustering step is computationally expensive, scaling quadratically with the number of input views, making it infeasible for high-resolution or large-scale scene processing.

To overcome these limitations, we propose FAST3DIS, a fully end-to-end framework for multi-view 3D instance segmentation that fundamentally eliminates the need for post-hoc clustering. Leveraging the powerful geometric priors of foundational depth models, we introduce a 3D-anchored, query-based Transformer architecture that formulates instance segmentation as a direct set prediction problem in continuous 3D space. Specifically, instead of grouping dense point-wise embeddings via expensive 3D heuristic clustering, our model directly decodes multi-view consistent masks from 2D observations. Our approach diverges from prior art through three core contributions:

\begin{itemize}
    \item 3D-Anchored Cross-Attention for Dense Segmentation: We formulate object queries as dynamic 3D anchors that serve as explicit spatial probes in continuous geometric space. We effectively extend sparse 3D detection paradigms~\cite{wang2022detr3d} to dense mask generation by projecting these 3D anchors into multi-view feature maps to sample local context via anchor-sampling cross-attention, which replaces dense global attention with deterministic 3D-to-2D query projections. This explicitly enforces multi-view geometric consistency while reducing the cross-attention complexity to scale linearly with the number of queries and views, thereby decoupling memory requirements from spatial resolution.
    
    \item Explicit Feature and Spatial Regularization: To achieve robust class-agnostic separation in a single forward pass, we introduce a dual-level regularization strategy. In the feature space, a multi-view contrastive loss pulls representations of the same 3D instance together across views while pushing distinct instances apart. In the spatial domain, we propose a dynamically scheduled overlap penalty that actively prevents distinct queries from claiming the same physical region, ensuring sharp instance boundaries without yielding artificial background gaps.
    
    \item Optimized Inference Efficiency and Scalability: By fundamentally bypassing the need for test-time heuristics and clustering, our approach significantly reduces the memory footprint and delivers orders-of-magnitude inference speedups without compromising competitive segmentation accuracy. This provides a highly scalable foundation for real-time 3D scene understanding.
\end{itemize}

\section{Related Work}

\subsection{Feed-Forward 3D Reconstruction}
The 3D reconstruction paradigm has recently shifted from multi-stage optimization (e.g., COLMAP~\cite{schoenberger2016sfm,schoenberger2016mvs}) to unified feed-forward Transformers~\cite{vaswani2017attention}. Methods like DUSt3R~\cite{wang2024dust3r}, Fast3R~\cite{2025Fast3R}, and FLARE~\cite{2025FLARE} demonstrated the ability to directly regress dense 3D pointmaps and camera parameters from unposed image collections. To overcome the batch-processing limitations of these offline methods, subsequent models such as CUT3R~\cite{2025CUT3R}, SLAM3R~\cite{2025SLAM3R}, and WINT3R~\cite{2025WINT3R} introduced incremental, real-time 3D reconstruction for streaming inputs.

Concurrently, visual geometry models, like VGGT~\cite{wang2025vggt}, FastVGGT~\cite{2025FastVGGT}, and Depth Anything V3 (DA3)~\cite{depthanything3}, have significantly advanced zero-shot metric depth estimation and spatial recovery. However, despite their robust geometric priors, these models lack semantic awareness and cannot distinguish object identities. Our framework addresses this gap by utilizing DA3 as a frozen geometric backbone. Through a LoRA~\cite{hu2022lora} adaptation, we extract instance-level semantics while fully preserving the model's generalized 3D reconstruction capabilities.

\subsection{Multi-View Instance Segmentation}
Early multi-view semantic approaches focus on fusing geometric and visual features. SAB3R~\cite{2025SAB3R} generates implicit multi-view semantic representations but lacks explicit consistency constraints. Alternatively, methods like Uni3R~\cite{2025Uni3R} and LSM~\cite{2024LSM} incorporate Vision-Language Models (VLMs, e.g., LSeg~\cite{2022LSeg}, CLIP~\cite{2021CLIP}) into 3D Gaussian Splatting~\cite{20233DGS}. While achieving cross-view semantic consistency, these VLM-driven representations are inherently category-level, failing to effectively discriminate between distinct instances of the same semantic class.

In the 2D domain, Mask2Former~\cite{cheng2021mask2former} established a unified mask classification paradigm by formulating segmentation as a set prediction problem. It utilizes learnable object queries, bipartite matching, and masked cross-attention to directly output mask probabilities. Extending 2D query-based segmentation to unposed multi-view scenarios introduces significant challenges in cross-view association. For example, PanSt3R~\cite{zust2025panst3r} jointly predicts geometry and panoptic masks based on MUSt3R~\cite{cabon2025must3r}. However, lacking explicit multi-view consistency constraints during training, PanSt3R optimizes masks primarily within individual 2D views. Consequently, it relies on a post-hoc quadratic binary optimization over the reconstructed point cloud to resolve cross-view conflicts, compromising its feed-forward efficiency.

To explicitly enforce multi-view consistency without test-time optimization, recent models like IGGT~\cite{li2025iggt} and UNITE~\cite{koch2025unified} adopt a post-hoc clustering paradigm. They learn a dense metric feature space via contrastive losses to group pixels belonging to the same instance, followed by non-differentiable density-based clustering (e.g., HDBSCAN~\cite{mcinnes2017accelerated}). While effective conceptually, this clustering step creates a severe latency and memory bottleneck. As the view count increases, dense point-wise clustering scales poorly, becoming computationally infeasible and highly sensitive to heuristic hyperparameters.

\subsection{Class-Agnostic 3D Instance Segmentation}
Traditional 3D instance segmentation approaches, such as Mask3D~\cite{2023Mask3D}, rely heavily on extensive point-level annotations. To bypass the need for massive 3D annotations, a prominent line of research leverages the zero-shot capabilities of 2D foundation models (e.g., SAM~\cite{2023SAM}, SAM2~\cite{2024SAM2}, CropFormer~\cite{2023CropFormer}). Training-free methods like SAM3D~\cite{2023SAM3D} and MaskClustering~\cite{2024MaskClustering} extract dense 2D masks from multi-view images and project them onto 3D point clouds via camera poses. Segment3D~\cite{2023segment3D} instead trains directly on SAM-generated pseudo-labels alongside limited ground truth, requiring only 3D point clouds during inference. Open3DIS~\cite{2024Open3DIS} partitions the 3D space into geometrically uniform superpoints to facilitate robust 2D-to-3D matching. Recent extensions, including Any3DIS~\cite{2025Any3DIS} and SAM2Object~\cite{2025SAM2Object}, incorporate SAM2's tracking capabilities to enhance multi-view temporal consistency. However, these 2D-lifting pipelines are fundamentally bottlenecked by severe multi-view occlusions and their reliance on hand-crafted, heuristic mask consolidation rules.

\subsection{3D-to-2D Query Projection}
To predict instance masks end-to-end across multiple views without any post-hoc optimization or spatial clustering, we draw inspiration from sparse 3D object detection frameworks (e.g., DETR3D~\cite{wang2022detr3d} and Deformable DETR~\cite{zhu2020deformable}). These methods pioneered the use of 3D-to-2D query projection, where 3D reference points are projected onto 2D camera views to sample local features for sparse bounding box regression. We adapt and extend this mechanism for dense 3D instance segmentation in unconstrained, unposed environments. We utilize learnable 3D anchors within the scene volume, coupled with anchor-sampling cross-attention to efficiently query features from different views by projecting 3D locations into 2D feature maps, bypassing the complexity of global attention.

\begin{figure*}[t]
    \centering
    \includegraphics[width=1.0\textwidth]{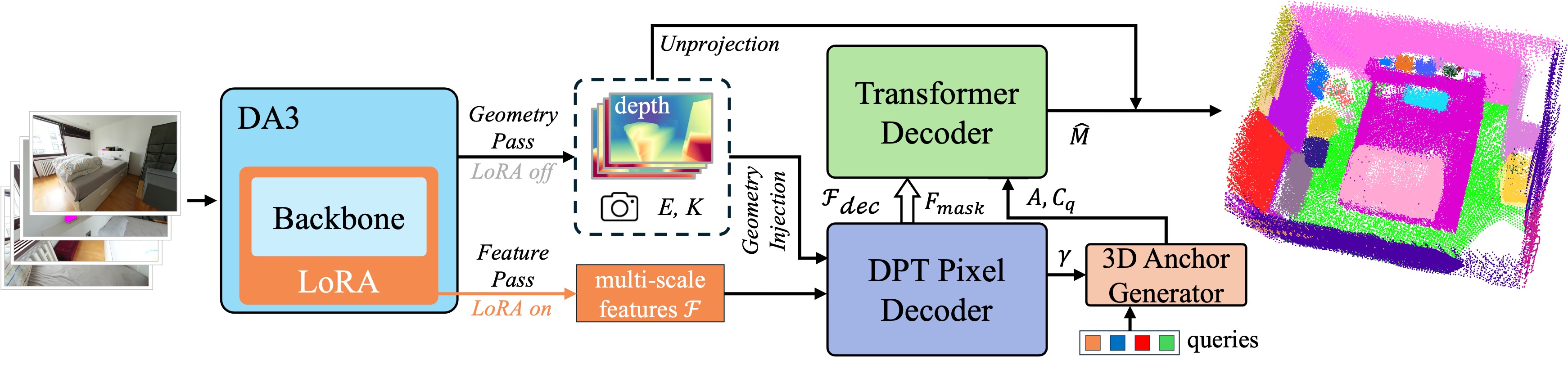}
    \caption{Overview of the FAST3DIS framework. Given unposed multi-view RGB images: (1) Dual-Pass Backbone: A frozen Depth Anything V3 (DA3)~\cite{depthanything3} extracts dense depth and camera parameters, while a LoRA~\cite{hu2022lora}-adapted pathway extracts multi-scale view features $\mathcal{F}$. (2) Geometry-Injected Pixel Decoder: The predicted geometry is injected to generate a 3D-aware feature pyramid $\mathcal{F}_{\text{dec}}$ and mask features $F_{\text{mask}}$. (3) 3D Anchor Generator: Explicit 3D anchors $A$ and content queries $C_q$ are produced conditioned on the global scene context $\gamma$. (4) Anchor-Sampling Transformer Decoder: 3D Anchors are onto the 2D views to sample local multi-view features from $\mathcal{F}_{\text{dec}}$. Finally, the predicted geometry and masks $\hat{M}$ are assembled into 3D instances.}
    \label{fig:overview}
\end{figure*}

\section{Method}
We propose FAST3DIS, an end-to-end feed-forward framework for multi-view 3D instance segmentation from unposed images, as illustrated in Figure~\ref{fig:overview}. Given a set of $S$ unposed RGB images $\mathcal{I} = \{I_1, \dots, I_S\}$ of shape $H \times W$, our model predicts a set of 3D-consistent binary masks and class identities without requiring intermediate depth supervision or post-hoc clustering. 

\subsection{Dual-Pass LoRA-Adapted Geometric Backbone}
To leverage strong geometric priors without the need for multi-view optimization, we adopt the pre-trained Depth Anything V3 (DA3)~\cite{depthanything3} as our foundational backbone. While DA3 excels at zero-shot metric depth and camera pose estimation, fine-tuning such a large-scale model for semantic instance segmentation often leads to catastrophic forgetting of its inherent 3D structural knowledge.

We propose a dual-pass adaptation strategy. We freeze the original DA3 parameters entirely and introduce Low-Rank Adaptation (LoRA)~\cite{hu2022lora} modules into the linear projections of DA3's backbone. During a forward pass, we execute two distinct pathways for each batch:
\begin{itemize}
    \item \textbf{Geometry Pass:} LoRA is disabled. We route the input images $\mathcal{I}$ through the frozen DA3 pipeline to extract dense metric depth maps $D$, along with the associated camera intrinsics $K$ and extrinsics $E$.
    \item \textbf{Feature Pass:} LoRA is enabled. We extract multi-scale view feature maps $\mathcal{F} = \{F^{1/32}, F^{1/16}, F^{1/8}, F^{1/4}\}$ from the DA3 backbone. The trainable LoRA parameters allow the backbone to adapt to instance-discriminative representations while implicitly sharing the underlying geometric feature space.
\end{itemize}

\subsection{Geometry-Injected Pixel Decoder}
To bridge the semantic features and the spatial 3D geometry, we employ a Dense Prediction Transformer (DPT)~\cite{ranftl2021vision} style pixel decoder. Standard pixel decoders operate purely in the 2D domain, which is insufficient for 3D consistency. Therefore, we explicitly inject the predicted 3D geometry into the decoding process.

Specifically, we unproject the depth map $D$ into a dense 3D point map $P_{\text{world}} \in \mathbb{R}^{S \times 3 \times H \times W}$ in the global coordinate system. This dense point map is downsampled to $1/32$ resolution via interpolation, projected to the embedding dimension, and directly added to the deepest backbone feature $F^{1/32}$. Following this geometry injection, a chain of feature fusion blocks progressively upsample and fuse the feature pyramid, and produce the refined multi-scale features $\mathcal{F}_{\text{dec}} = \{F_{\text{dec}}^{1/32}, F_{\text{dec}}^{1/16}, F_{\text{dec}}^{1/8}\}$ for the Transformer decoder, alongside a high-resolution mask feature $F_{\text{mask}} \in \mathbb{R}^{S \times C \times \frac{H}{4} \times \frac{W}{4}}$. To ensure cross-view stability, $F_{\text{mask}}$ is explicitly regularized by a geometric consistency loss (Section~\ref{sec:loss}).

\subsection{Learned 3D Anchor Generator}
Extending standard query-based 2D segmentation models (e.g., Mask2Former~\cite{cheng2021mask2former}) to 3D might struggle to consistently associate the same physical object across dramatically different viewpoints. To physically ground the queries, we introduce a learned 3D anchor generator. We define $N_q$ learnable 3D anchor points $A_{init} \in \mathbb{R}^{N_q \times 3}$, and employ a multi-layer perceptron (MLP) to predict a dynamic 3D shift $\Delta A$ conditioned on the global scene context $\gamma$ obtained via global average pooling of the deepest feature $F_{1/32}$. The refined anchors are computed as:
\begin{equation}
    A = A_{init} + \text{MLP}(\gamma)
\end{equation}
Concurrently, the generator outputs $N_q$ content queries $C_q \in \mathbb{R}^{N_q \times C}$. These 3D anchors act as explicit spatial priors, representing the physical centers of potential objects in the continuous world coordinate system.

\subsection{Anchor-Sampling Cross-Attention Decoder}
The core bottleneck in multi-view attention mechanisms is the sequence length. Standard cross-attention between $N_q$ queries and all pixels across $S$ views yields a prohibitive computational complexity of $\mathcal{O}(N_q \cdot S \cdot HW)$. Building upon the multi-scale paradigm of Mask2Former~\cite{cheng2021mask2former}, we employ an anchor-sampling cross-attention that bypasses global dense matching, reducing the spatial complexity from $\mathcal{O}(HW)$ to $\mathcal{O}(1)$.

Our Transformer decoder consists of $L$ layers. Following Mask2Former, we adopt a round-robin routing strategy for the multi-scale features: at each decoder layer $l$, the cross-attention operates on a specific feature scale $F_{\text{dec}}^m \in \mathcal{F}_{\text{dec}}$. 

Let $C_q^{(l-1)}$ and $A^{(l-1)}$ denote the content queries and the corresponding 3D anchors by the previous layer (or initialized by the generator when $l=1$). Inside the $l$-th decoder layer, the queries are processed through three main steps:

\textbf{1. Query Self-Attention:} We first inject the 3D spatial priors into the content queries using a Fourier-based 3D positional encoding ($\text{PE}_{\text{3D}}$):
\begin{equation}
    q_i = C_{q, i}^{(l-1)} + \text{PE}_{\text{3D}}(A_i^{(l-1)})
\end{equation}
These queries are then processed via a Multi-Head Self-Attention (MHSA)~\cite{vaswani2017attention} module, interacting with each other to reason about global scene context.

\textbf{2. Anchor-Sampling Cross-Attention:} Instead of globally attending to the dense feature map $F_{\text{dec}}^m$, each updated query $q_i$ explicitly leverages its corresponding 3D anchor $A_i^{(l-1)}$. For each view $s \in \{1, \dots, S\}$, we project the 3D anchor onto the 2D view using the predicted extrinsics $E_s$ and intrinsics $K_s$:
\begin{equation}
    [u_{i,s}, v_{i,s}, z_{i,s}]^\top = K_s \left( E_s \cdot [A_i^{(l-1)}, 1]^\top \right)
\end{equation}
The normalized 2D coordinates $(u_{i,s}/z_{i,s}, v_{i,s}/z_{i,s})$ are then used to bilinearly sample the chosen feature map $F^{\text{dec}}_m$, yielding a view-specific local feature vector $f_{i,s}^{(l)} \in \mathbb{R}^C$. The cross-attention for the $i$-th query is then performed exclusively over its $S$ spatially sampled features:
\begin{equation}
    \text{CrossAttn}(q_i, \mathcal{K}, \mathcal{V}) \quad \text{where} \quad 
    \mathcal{K} = \mathcal{V} = \{f_{i,1}^{(l)}, f_{i,2}^{(l)}, \dots, f_{i,S}^{(l)}\}
\end{equation}
This forces the query to aggregate visual evidence from the same physical 3D location across all views, explicitly enforcing multi-view geometric consistency.

\textbf{3. Iterative Anchor Refinement:} After a Feed-Forward Network (FFN) updates the content query to $C_q^{(l)}$, we apply an iterative refinement to the 3D anchors. We pass the updated content query through a lightweight MLP to predict a residual 3D shift $\Delta A^{(l)}$, allowing the model to progressively adjust the physical location of the anchor to better center on the target object:
\begin{equation}
    A_i^{(l)} = A_i^{(l-1)} + \text{MLP}(C_q^{(l)})
\end{equation}

After $L$ layers of progressive semantic refinement and spatial adjustment, the final outputs are highly discriminative query embeddings and precisely localized 3D instance centers, which are then routed to the mask generation module.

\subsection{Multi-View Mask Generation and 3D Assembly}
The Transformer decoder ultimately produces $N_q$ updated query embeddings. These embeddings are passed through two parallel branches: a classification head that predicts the objectness (or class logits) $p \in \mathbb{R}^{N_q \times 2}$, and an MLP that generates mask embeddings $\mathcal{E}_{\text{mask}} \in \mathbb{R}^{N_q \times C}$.

To obtain the final 2D instance masks for each view, we expand the mask embeddings $\mathcal{E}_{\text{mask}}$ to shape $S \times N_q \times C$, and compute the dot product between these embeddings and the high-resolution per-view pixel features $F_{\text{mask}}$:
\begin{equation}
    \hat{M}_{s} = \sigma \left( \mathcal{E}_{\text{mask}} \cdot F_{\text{mask}}^{(s)} \right) \in \mathbb{R}^{N_q \times \frac{H}{4} \times \frac{W}{4}}
\end{equation}
where $\sigma$ denotes the sigmoid activation function, and $\hat{M}_{s}$ represents the multi-view consistent 2D binary masks for the $s$-th view.

During inference, for each query $i$ with an objectness score above a predefined threshold $\theta$, we collect all valid pixels $(u, v)$ across all views where the predicted mask probability $\hat{M}_{s}^{(i, u, v)} > \theta$. Utilizing the purely feed-forward metric depth $D_s$ and camera parameters $K_s, E_s$ from our geometry pass, we directly unproject these pixels into the 3D world space. The union of these unprojected points across all views inherently forms the dense, complete 3D point cloud of the $i$-th instance, achieving unified 3D instance segmentation in a single forward pass.

\subsection{Training Objectives} \label{sec:loss}
Our framework is trained end-to-end following the set prediction paradigm. We first establish an optimal one-to-one bipartite matching between the $N_q$ predictions and the $N_{\text{gt}}$ ground-truth objects using the Hungarian algorithm~\cite{kuhn1955hungarian}. Since FAST3DIS performs class-agnostic instance segmentation, the matching cost only considers the objectness probability and the geometric mask similarity. Once the optimal assignment is determined, the network is optimized using a unified objective encompassing standard 2D segmentation losses, explicit 3D feature consistency, and a dynamic boundary overlap penalty.

\paragraph{Standard Segmentation Losses.} For the matched queries, we utilize a combination of traditional instance segmentation losses. To handle the foreground-background class imbalance and optimize the class-agnostic objectness score, we apply the Sigmoid Focal Loss ($\mathcal{L}_{\text{cls}}$). For the dense mask prediction, we employ Binary Cross-Entropy (BCE) loss ($\mathcal{L}_{\text{bce}}$) for pixel-wise classification, paired with the Sørensen–Dice loss ($\mathcal{L}_{\text{dice}}$) to enforce global shape agreement and mitigate the impact of object scale variations.

\paragraph{View-Consistent Contrastive Optimization.} 
While standard segmentation losses operate independently on individual 2D views, they lack the cross-view geometric constraints necessary for 3D segmentation. We explicitly regularize the underlying dense mask feature space via a point-level supervised contrastive loss~\cite{khosla2020supervised}. Specifically, we perform instance-balanced sampling across the multi-view feature maps to construct a dense subset of points $I$, where each point $i \in I$ is mapped to its ground-truth instance identity $y_i$. For a point $i$, let $z_i$ denote its $L_2$-normalized feature, and let $P(i) \equiv \{p \in I \setminus \{i\} : y_p = y_i\}$ be the set of its positive matches. The multi-view geometric consistency loss is formulated as:
\begin{equation}
    \mathcal{L}_{\text{geo}} = \sum_{i \in I} \frac{-1}{|P(i)|} \sum_{p \in P(i)} \log \frac{\exp(z_i \cdot z_p / \tau)}{\sum_{a \in I \setminus \{i\}} \exp(z_i \cdot z_a / \tau)}
\end{equation}
where $\tau$ is a temperature hyperparameter. By optimizing densely sampled pixel features, this formulation yields a discriminative feature space directly optimized for precise boundary separation.

\paragraph{Dynamic Spatial Overlap Penalty.} During preliminary experiments, we observed a frequent failure mode where multiple queries would "hijack" or compete for the same physical instance across views. Relying solely on standard segmentation losses with Hungarian matching on the more complex 3D-consistent class-agnostic segmentation is challenging. Without class supervision to penalize queries for merging adjacent objects, the model lacks a critical natural regularizer. For example, while a query may perfectly segment an object in one view, occlusions can cause its mask to inappropriately expand in alternate views. Global bipartite matching cannot provide the strict, pixel-level cross-view exclusivity required here, inevitably leading to overlapping masks and semantic bleeding. Furthermore, while our contrastive loss $\mathcal{L}_{\text{geo}}$ enforces instance distinction in the feature space, its sample-based nature cannot guarantee the defense of hard spatial boundaries at the pixel level. 

To explicitly penalize active queries for claiming the same spatial region, we introduce a confidence-weighted pairwise overlap loss. Let $\mathbf{p}_i \in [0, 1]^M$ denote the flattened multi-view spatial probability map for the $i$-th query, and $c_i$ denote its predicted objectness score. The loss is computed as the sum of weighted dot products between all distinct query pairs:
\begin{equation}
\mathcal{L}_{\text{overlap}} = \frac{1}{N_q(N_q-1)} \sum_{i=1}^{N_q} \sum_{j \neq i} (c_i \mathbf{p}_i) \cdot (c_j \mathbf{p}_j)
\end{equation}

However, naively applying a static, strong overlap penalty creates a suboptimal optimization landscape. If the penalty is too aggressive during early training, it encourages ``lazy'' queries that universally predict low confidence scores and are thus filtered into the background class, safely avoiding overlap collisions. Conversely, maintaining a high penalty during the training encourages the network to predict artificial background safety zones between adjacent instances. 

To resolve this, we dynamically schedule the overlap weight $\lambda_{\text{overlap}}(t)$ over the normalized training progress $t \in [0, 1]$. Initially ($t \le 0.6$), $\lambda_{\text{overlap}}$ exponentially warms up to its maximum, allowing queries to freely explore without large penalties. Subsequently, it linearly decays to a minimal floor, ensuring tight convergence on object boundaries to reduce artificial background gaps.

\paragraph{Overall Objective.} To facilitate deep supervision, the primary segmentation and overlap losses are applied to the outputs of all intermediate Transformer decoder layers. The final training objective is formulated as:
\begin{equation}
    \mathcal{L}_{\text{total}} = \sum_{l=1}^{L} \Big( \lambda_{\text{cls}}\mathcal{L}_{\text{cls}}^{(l)} + \lambda_{\text{bce}}\mathcal{L}_{\text{bce}}^{(l)} + \lambda_{\text{dice}}\mathcal{L}_{\text{dice}}^{(l)} + \lambda_{\text{overlap}}(t)\mathcal{L}_{\text{overlap}}^{(l)} \Big) + \lambda_{\text{geo}}\mathcal{L}_{\text{geo}}
\end{equation}

\section{Experiments}

\subsection{Experimental Setup}

\paragraph{Implementation Details.} All evaluations are conducted on a computing node equipped with a 32-core AMD EPYC 9654 processor and a single NVIDIA H200 GPU (144GB VRAM). Both our FAST3DIS and IGGT resize all input images to a long-edge resolution of 504 pixels, following DA3 and VGGT standard.

\paragraph{Datasets.} Our model is trained exclusively on the Aria Synthetic Environments Dataset~\cite{pan2023aria}, which provides highly photorealistic simulated indoor environments with rich multi-view multi-modal annotations. Due to the computational overhead considering the size of the full dataset ($100,000+$ scenes), we sampled $40\%$ of the scenes to form our training set. We conducted a statistical analysis of the ground-truth instance distributions across the training scenes and set the number of learnable 3D anchors and content queries in our model to $N_q = 80$. 

To assess the zero-shot generalization and geometric robustness, our evaluations are on three highly diverse, standard 3D scene understanding datasets:
(1)ScanNet V2~\cite{dai2017scannet}: A widely adopted, large-scale dataset containing complex, real-world indoor scenes. (2) ScanNet++~\cite{yeshwanth2023scannet++}: A recent benchmark that provides high-resolution indoor scenes captured with sub-millimeter laser scanners. (3) Replica~\cite{straub2019replica}: A dataset consisting of highly photorealistic synthetic indoor environments with high-fidelity geometry and dense semantic annotations.

\paragraph{Baselines.} Conventional 3D instance segmentation methods operating on pre-reconstructed, high-fidelity point clouds or posed RGB-D sequences may achieve higher absolute ``segmentation scores''. However, such methods impose strict data and preprocessing dependencies. In contrast, our method requires only unposed multi-view RGB images, harnessing the powerful geometric priors of the DA3 backbones to simultaneously recover 3D geometry and instance boundaries.

Thus, the primary objective of our evaluation is to compare against  recent state-of-the-art feed-forward 3D scene understanding models. Despite the fact that PanSt3R~\cite{zust2025panst3r} shares similar geometric foundation, we do not focus our primary comparison on it due to its divergent task formulation (panoptic segmentation) and its reliance on test-time optimization, as previously discussed. Instead, we explicitly position our method against IGGT~\cite{li2025iggt}, which represents the post-hoc clustering paradigm for class-agnostic 3D instance segmentation. We aim to demonstrate that our purely end-to-end architecture structurally resolves the severe latency and memory bottlenecks inherent in clustering, while maintaining highly competitive segmentation accuracy.

Despite this positioning, our quantitative evaluation on class-agnostic 3D instance segmentation includes a broader comparison. Specifically, we compare with : (1) \textit{Geometry-only clustering methods} HDBSCAN~\cite{mcinnes2017accelerated} and Felzenszwalb et al.~\cite{felzenszwalb2004efficient} relying entirely on spatial coordinates; (2) \textit{2D-driven models} SAM3D~\cite{2023SAM3D} and Segment3D~\cite{2023segment3D}, which leverage the 2D segmentation capabilities of SAM~\cite{2023SAM} and project them into 3D space; and (3) PanSt3R and IGGT.

\subsection{Efficiency and Scalability Analysis}
The key advantage of FAST3DIS over IGGT lies in its end-to-end inference efficiency and memory scalability. While IGGT utilizes efficient geometric backbones VGGT, their overall inference pipeline is heavily bottlenecked by the post-hoc clustering stage. In the post-hoc clustering paradigm, algorithms like HDBSCAN require constructing dense, high-dimensional pairwise distance matrices across all point embeddings. When attempting to accelerate this process using GPU-optimized implementations (e.g., cuML's GPU HDBSCAN), the $\mathcal{O}(N^2)$ memory footprint quickly becomes a fatal flaw. In our testing environment, IGGT equipped with GPU HDBSCAN triggers an Out-Of-Memory (OOM) error when attempting to process images. However, real-world scene reconstruction and segmentation often necessitate significantly more dense view coverage. To process larger numbers of views without hitting the VRAM ceiling, one is forced to fallback to the CPU implementation. While this avoids OOM errors, the extreme latency makes it unscalable.
 
\begin{figure*}[t]
    \centering
    \includegraphics[width=1.0\textwidth]{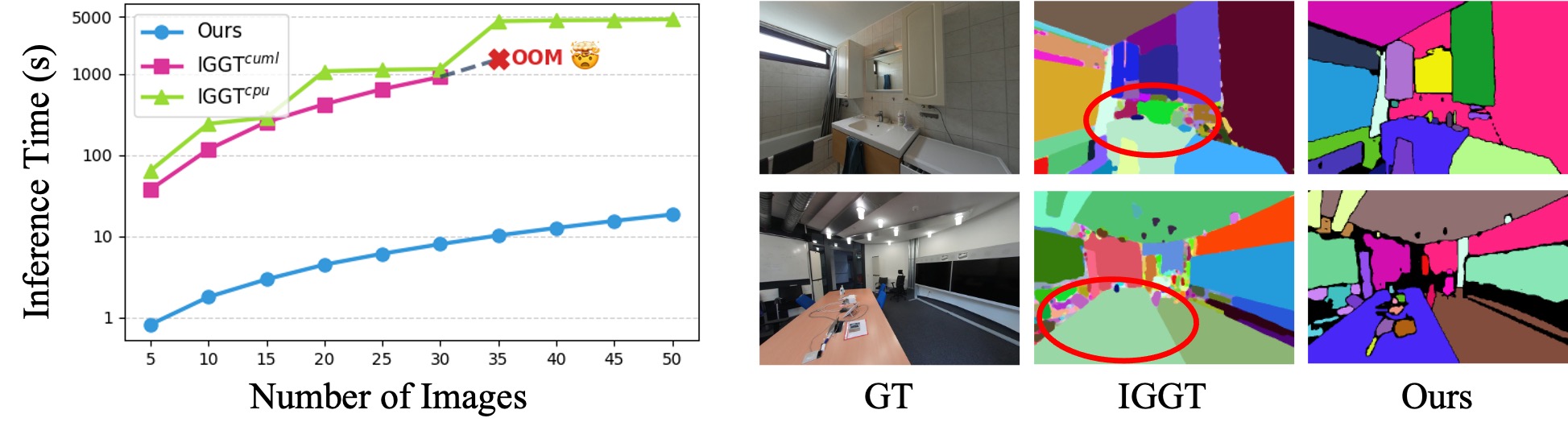}
    \caption{Comparisons with IGGT~\cite{li2025iggt}. \textbf{Left:} Inference time comparison (log scale). \textbf{Right:} Qualitative segmentation results. Red circles highlight small objects that IGGT fails to isolate and incorrectly merges into the supporting background.}
    \label{fig:comparison}
\end{figure*}

As illustrated in ~\autoref{fig:comparison}, the latency gap between FAST3DIS and IGGT widens exponentially as the number of views increases. At 30 input views, the inference time for FAST3DIS is 7.9 seconds. In contrast, IGGT requires 910.8 seconds with GPU HDBSCAN, and 1141.5 seconds using the CPU version. Our method achieves over a $115\times$ speedup. At 50 input views, while IGGT on GPU fails entirely due to OOM, IGGT on CPU takes approximately 4663.3 seconds to process a single scene. In comparison, FAST3DIS smoothly handles the same 50-view input in 18.4 seconds, representing a $250\times$ acceleration. Our experiments show that the additional VRAM overhead introduced by our method is marginal compared to the frozen DA3 backbone itself. We evaluated up to 350 images (682.7 seconds) in a single forward pass without approaching the hardware limits, demonstrating that our architecture is structurally equipped for large-scale, unconstrained multi-view environments.

\subsection{Qualitative Evaluation}
\begin{figure*}[t]
    \centering
    \includegraphics[width=1.0\textwidth]{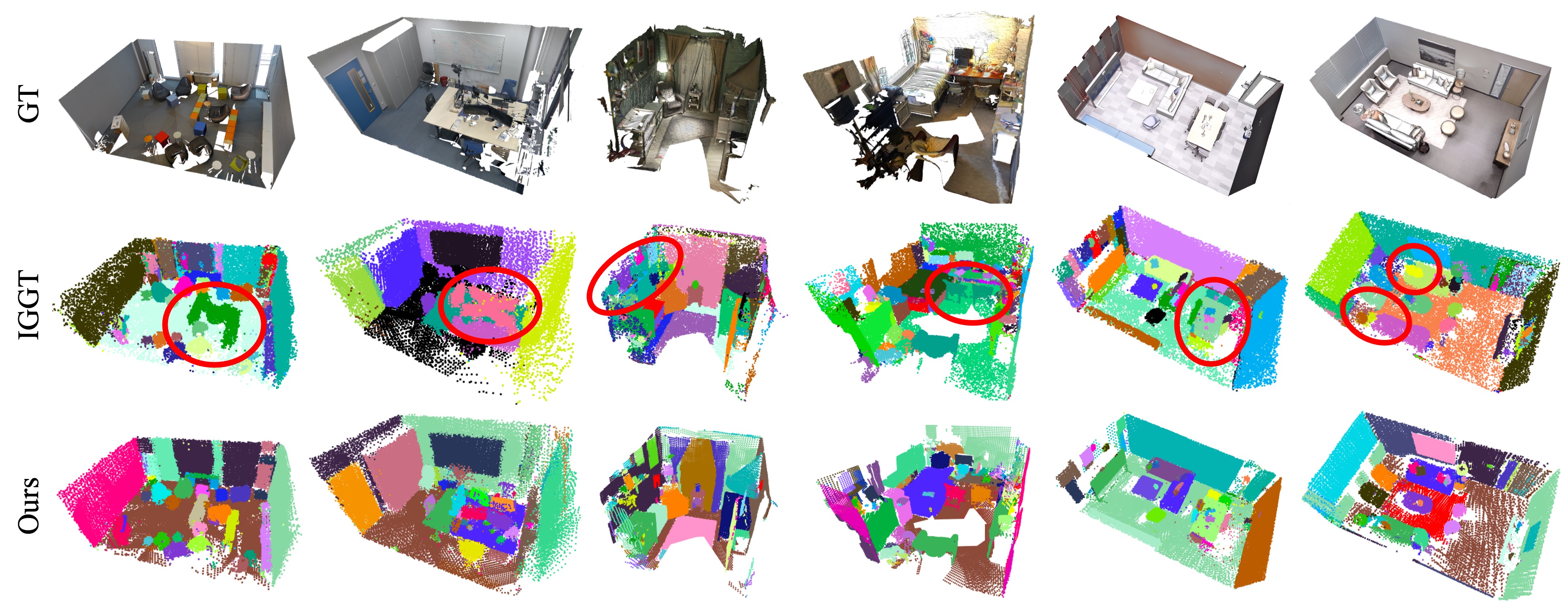}
    \caption{Qualitative comparison with IGGT. Red circles highlight adjacent objects that IGGT incorrectly merges into a single instance.}
    \label{fig:qualitative}
\end{figure*}

To gain deeper insights into the structural advantages of our end-to-end architecture, we qualitatively compare the segmentation outputs of FAST3DIS against IGGT. As discussed, IGGT relies on HDBSCAN to group 3D points. In contrast, FAST3DIS predicts masks directly via physically distinct 3D anchors. This architectural difference is evident in two challenging scenarios:

\begin{itemize}
    \item \textbf{Segmentation of Small Objects.} A critical vulnerability of clustering-based method is the scale and density imbalances. Figure~\ref{fig:comparison} shows example 2D segmentation masks, highlighting scenarios where small objects rest on large supporting structures (e.g., small items on a table and on a bathroom sink). In IGGT's feature space, these small objects generate relatively few 3D points compared to the massive supporting surfaces. HDBSCAN might cause them to be "swallowed" and merged into the background object. In contrast, FAST3DIS is more sensitive to small objects.
    \item \textbf{Separation of Adjacent Objects.} Another major failure mode of clustering occurs when distinct objects of similar semantics are physically adjacent or touching. As shown in Figure~\ref{fig:qualitative}, IGGT sometimes struggles with "semantic bleeding". Because the contrastive loss pulls similar features together, and the objects share a continuous physical boundary, the local geometric density remains uninterrupted. Consequently, HDBSCAN merges the adjacent objects into a single instance. In comparison, FAST3DIS's queries compete for exclusive spatial occupancy and better locate the boundaries between adjacent instances, driven by 3D anchors and the spatial overlap penalty.
\end{itemize}

\subsection{Quantitative Evaluation}

To quantitatively evaluate the performance of our proposed method, we employ the standard Average Precision (AP) metrics widely adopted for 3D instance segmentation tasks. In the class-agnostic setting, we ignore the semantic class labels in the annotations and focus purely on object localization and boundary quality. We report the AP at Intersection-over-Union (IoU) thresholds of 25\% ($\text{AP}_{25}$) and 50\% ($\text{AP}_{50}$), as well as the overall $\text{AP}$, which is averaged across IoU thresholds ranging from 50\% to 95\% with a step size of 5\%. 

Unlike other baselines that operate directly on pre-aligned ground-truth geometries, feed-forward 3D reconstruction models (FAST3DIS, IGGT, PanSt3R) predict point clouds and camera poses in a first-view relative, unscaled local coordinate system. We apply a Sim(3) trajectory alignment followed by an ICP refinement to align their predictions for evaluation. Correspondences are established by mapping each predicted point to its nearest ground-truth vertex. To ensure a fair and computationally feasible comparison, due to the GPU memory and latency limitations faced by IGGT as discussed, we uniformly sample 50 views along the camera trajectory to reconstruct and evaluate each scene.

Quantitative results are reported in Table~\ref{table:quantitative}. On ScanNet V2 and Replica, FAST3DIS demonstrates competitive performance, outperforming our primary baseline IGGT, and achieves comparable results to 2D-driven foundation models. 

\begin{table*}[t]
\scriptsize
\centering
\caption{\textbf{3D Instance Segmentation Evaluation on ScanNet, ScanNet++, and Replica.} We report standard Average Precision (AP) metrics. The best results are highlighted in \textbf{bold}, and the second-best results are \underline{underlined}.}
\begin{tabular}{l ccc ccc ccc}
\toprule
\multirow{2}{*}{\textbf{Method}} & \multicolumn{3}{c}{\textbf{ScanNet V2~\cite{dai2017scannet}}} & \multicolumn{3}{c}{\textbf{ScanNet++~\cite{yeshwanth2023scannet++}}} & \multicolumn{3}{c}{\textbf{Replica~\cite{straub2019replica}}} \\
\cmidrule(lr){2-4} \cmidrule(lr){5-7} \cmidrule(lr){8-10}
& AP$_{25}$ & AP$_{50}$ & AP & AP$_{25}$ & AP$_{50}$ & AP & AP$_{25}$ & AP$_{50}$ & AP \\
\midrule
HDBSCAN~\cite{mcinnes2017accelerated} & 5.6 & 0.6 & 0.2 & 2.4 & 0.6 & 0.2 & 2.3 & 0.1 & 0.0 \\
Felzenszwalb~\cite{felzenszwalb2004efficient} & 10.0 & 2.2 & 0.8 & 2.9 & 1.1 & 0.5 & 1.6 & 1.4 & 1.0 \\
\midrule
SAM3D~\cite{2023SAM3D} & 23.3 & 8.0 & 2.4 & \underline{19.2} & \underline{7.0} & \underline{2.2} & \underline{28.3} & \textbf{16.1} & \textbf{7.1} \\
Segment3D~\cite{2023segment3D} & 17.2 & 7.2 & 2.5 & 17.3 & \textbf{8.1} & \textbf{3.1} & 15.8 & 9.3 & 4.0 \\
\midrule
PanSt3R~\cite{zust2025panst3r} & 22.5 & 3.5 & 0.8 & \textbf{21.7} & 6.2 & 1.7 & 20.7 & 6.2 & 1.7 \\
IGGT~\cite{li2025iggt} & \underline{28.7} & \textbf{11.2} & \underline{2.8} & 10.0 & 2.2 & 0.6 & 15.2 & 3.9 & 1.1 \\
\textbf{FAST3DIS} (Ours) & \textbf{31.6} & \underline{9.6} & \textbf{3.8} & 14.7 & 3.5 & 1.0 & \textbf{30.3} & \underline{13.8} & \underline{5.1} \\
\bottomrule
\end{tabular}
\label{table:quantitative}
\end{table*}

\textbf{Insight on Highly Cluttered Scenes:} It is worth noting the general performance degradation observed for both FAST3DIS and IGGT on the ScanNet++ dataset. Through empirical analysis, we attribute this to a fundamental capacity bottleneck when facing dense annotations. ScanNet++ scenes are exhaustively annotated, frequently containing upwards of 100 to 200 ground-truth instances per scene. For FAST3DIS, our Transformer decoder was intentionally configured with $N_q = 80$ queries based on the statistical distribution of our training data, causing the network to naturally miss objects in hyperscale scenes. Interestingly, while IGGT is not bounded by a fixed query count, its performance also drops sharply. Our analysis reveals that clustering-based pipelines suffer from severe under-segmentation in these highly cluttered environments. Due to semantic bleeding, IGGT might merges distinct objects into single large clusters, typically predicting only 20 to 30 instances per Scannet++ scene.

\subsection{Ablation Studies}
\begin{figure*}[t]
    \centering
    \includegraphics[width=1.0\textwidth]{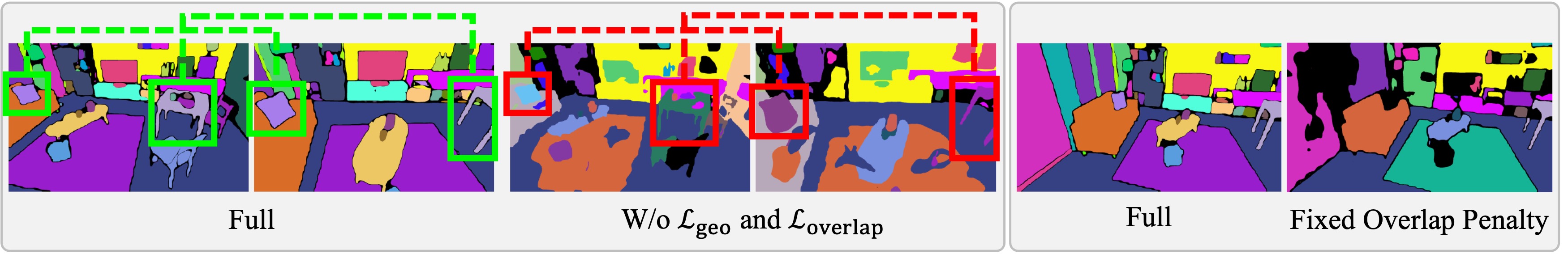}
    \caption{Qualitative ablation of our proposed regularization strategies. \textbf{Left:} Impact of explicit feature and spatial regularization. The full model successfully maintains consistent instance identities across different viewpoints (highlighted in green). Without $\mathcal{L}_{\text{geo}}$ and $\mathcal{L}_{\text{overlap}}$ (see Section~\ref{sec:loss}), the model suffers from severe cross-view association failures (the table and the pillow, highlighted in read) and ``query hijacking'', resulting in physically overlapping masks. \textbf{Right:} Impact of the dynamic overlap penalty. A fixed, static penalty forces the network to conservatively predict artificial background gaps (black regions) between adjacent objects. In contrast, our dynamic scheduling ensures more precise and contiguous instance boundaries.}
    \label{fig:ablation}
\end{figure*}

To validate the specific optimization designs of FAST3DIS, we conduct ablation studies focusing explicitly on our custom loss formulations designed for class-agnostic multi-view segmentation. ~\autoref{fig:ablation} illustrates the following cases:

\begin{itemize}
    \item \textbf{Necessity of Feature and Spatial Regularization.} A natural question is whether existing feed-forward panoptic models (e.g., PanSt3R) can be directly repurposed for class-agnostic instance segmentation by simply discarding their semantic classification branches. To simulate this naive adaptation, we ablate both our view-consistent contrastive loss ($\mathcal{L}_{\text{geo}}$) and the spatial overlap penalty ($\mathcal{L}_{\text{overlap}}$). This effectively reduces our training objective to a standard multi-view Mask2Former~\cite{cheng2021mask2former} formulation. 

    The empirical results reveal a severe degradation in mask quality and instance separation. Without explicit 3D feature regularization and spatial anti-collision constraints, the network exhibits frequent ``query hijacking'' where multiple queries converge on and claim the exact same physical object. Furthermore, the predicted multi-view masks suffer from highly ambiguous and blurry boundaries. This proves that while standard bipartite matching is sufficient for panoptic tasks (where distinct semantic labels naturally pull adjacent objects apart), pure class-agnostic 3D segmentation fundamentally requires additional spatial regularization to extract object boundaries.

    \item \textbf{Fixed vs. Dynamic Overlap Penalty.} Having established the absolute necessity of the overlap penalty to prevent ``query hijacking'', we analyze the impact of its weighting strategy. Applying a consistently high static weight for $\mathcal{L}_{\text{overlap}}$ helps to force queries to claim disjoint regions, but meanwhile introduces another failure mode: The network becomes overly conservative, predicting thick "black gaps" that are artificial background safety zones due to low confidence of all queries, between adjacent instances to safely avoid any penalty collision. Conversely, maintaining a consistently low overlap weight fails to fully resolve the initial hijacking problem. Our proposed dynamic weight scheduling directly addresses this dilemma. By exponentially warming up the penalty during early training to separate queries, and linearly decaying it in the final stages, the model is encouraged to confidently extend predictions to the true object edges.
\end{itemize}

\section{Conclusion}
In this paper, we presented FAST3DIS, a fully end-to-end framework for multi-view 3D instance segmentation that fundamentally overcomes the severe latency and memory bottlenecks of traditional post-hoc clustering paradigms. By incorporating a learned 3D anchor generator coupled with an efficient anchor-sampling cross-attention mechanism, our approach avoids the quadratic complexity of global attention and explicitly enforces multi-view geometric consistency. Furthermore, our dual-level regularization strategy, which integrates multi-view contrastive learning with a dynamically scheduled spatial overlap penalty to effectively prevents query collisions and ensures precise boundary. Experiments demonstrate that FAST3DIS not only achieves highly competitive segmentation accuracy within the pure feed-forward 3D paradigm on standard benchmarks, but also delivers orders-of-magnitude inference speedups over clustering-based methods, particularly as the number of input views scales.

%
%
\bibliographystyle{splncs04}
\bibliography{main}
\end{document}